\documentclass{article}
\usepackage{ijcai20}

\usepackage{times}
\usepackage{soul}
\usepackage{url}
\usepackage[hidelinks]{hyperref}
\usepackage[utf8]{inputenc}
\usepackage[small]{caption}
\usepackage{graphicx}
\usepackage{amsmath}
\usepackage{amsthm}
\usepackage{booktabs}
\usepackage{algorithm}
\usepackage{algorithmic}
\usepackage{graphicx}
\usepackage{t1enc}
\usepackage{multirow}
\usepackage{array} 
\usepackage{array,hhline} %To create tables and matrices.
\usepackage{rotating} %To rotate a table.
\usepackage{tabularx} %An extended version of tabular
\usepackage{amsmath}
\newcommand{\mbeq}{\overset{!}{=}}

\title{Real-Time Lane ID Estimation Using Recurrent Neural Networks With Dual Convention}

\author{
Ibrahim Halfaoui$^1$
\and
Fahd Bouzaraa$^1$\and
Onay Urfalioglu$^{^1}$\And
Li Minzhen$^2$
\affiliations
$^1$ Huawei Munich Research Center, Germany\\
$^2$ Huawei Intelligent Automotive Solution BU, China \\
\emails
\{ibrahim.halfaoui, fahd.bouzaraa, onay.urfalioglu, liminzhen\}@huawei.com,
}

\begin{document}

\maketitle

\begin{abstract}  % put your abstract here!
Acquiring information about the road lane structure is a crucial step for autonomous navigation.
To this end, several approaches tackle this task from different perspectives such as lane marking detection or semantic lane segmentation. 
However, to the best of our knowledge, there is yet no purely vision based end-to-end solution to answer the precise question: How to estimate the relative number or "ID" of the current driven lane within a multi-lane road or a highway? \\
In this work, we propose a real-time, vision-only (i.e. monocular camera) solution to the problem based on a dual left-right convention. We interpret this task as a classification problem by limiting the maximum number of lane candidates to eight. 
%(very unlikely to have a road with more than this number of lanes).
%Typical challanges for end-to-end lane ID estimation are occlusions by large vehicles as well as multi-lane scenes, where 
%lane markings are not fully visible due to dense traffic.
%We obtain a large dataset from various areas, containing labeled driving sequences, each ranging multiple kilometers.
Our approach is designed to meet low-complexity specifications and limited runtime requirements. It harnesses the temporal dimension inherent to the input sequences to improve upon high-complexity state-of-the-art models. We achieve more than 95\% accuracy on a challenging test set with extreme conditions and different routes. 
\end{abstract}

\section{Introduction}
As modern autonomous driving systems are progressively targeting the general consumer market, they have become increasingly reliant on the analysis and processing of visual information provided by mounted cameras, which represent for many applications a suitable alternative to expensive sensors such as high accuracy GPS or LIDARs. 
This was mostly enabled by the remarkable advancement of visual-based approaches in robustness and accuracy, which obviously benefited from the recent progress in matters of artificial intelligence and machine learning. \\
More specifically, self-driving cars operating in level three or higher require an accurate representation and understanding of the surrounding environment as a pre-requisite for efficient decision-making and safe drive control. For this, many tasks prove necessary including pixel-level semantic segmentation, object recognition \& detection, mapping \& localization and safe path planing, among others. For the latter two, an accurate lane-level knowledge about the position of the car in a multi-lane road or a highway could play a central role in improving the accuracy of the outcome of these applications. This revolves around the capability of the car to localize itself on lane-level and determine on which lane it is currently driving relative to a fixed reference such as the left or right borders of the road. The resulting knowledge represents valuable features which can be stored and added to the map. Similarly, it can contribute to improving the localization step and the planing for the safest path to take. \\
\section{Related Work}
%The recent rise of interest around artificial intelligence topics such as autonomous driving or robot navigation resulted in a new set of open research topics which are crucial to these technologies. Accordingly, understanding and analyzing the environment surrounding the robot or the autonomous vehicle is a key component to assure proper navigation. 
Knowing exactly the current lane of the vehicle on a multi-lane road could be necessary for many applications. This procedure is known as 'Lane ID estimation' not to be confused with the task of lane detection, segmentation or estimation that is widely popular in the literature and defined as the task of semantically distinguishing the drivable lane under the form of a distinguished pixel-level labeled entity. 
A notable approach for accurately estimating the ID of the current driven lane is the LaneQuest~\cite{lanequest} approach. It is a technique that relies on ubiquitous inertial sensors available in commodity smart-phones to provide an accurate estimate of the car's current lane without any visual input.
%but rather based on information about the surrounding environment of the car.
In~\cite{gpsppp}, a GPS-based method known as GPS-PPP is introduced. It allows for sub-meter real-time accurate positioning of vehicles on multi-lane motorways. 
~\cite{intvec_com} suggests a Markov-based alternative for lane-level localization that leverages connectivity between neighboring cars taking part in the traffic within a certain range and exchanging information to precisely locate each other.
%All the aforementioned methods rely either on GPS, inertial or inter-vehicle shared information to precisely estimate the current driven lane but do not consider any visual input information (images/video) to perform lane ID estimation.  
There exist, however, certain approaches which harness visual cues to perform the lane-level positioning for autonomous vehicles.
~\cite{stereov} Proposes an accurate real-time positioning method for robotic cars in urban environments. This method uses a robust lane marking detection algorithm, as well as an efficient shape registration algorithm between the detected lane markings and a GPS-based road shape prior, to improve the robustness and accuracy of the global localization.
In~\cite{3dlane}, a 3D lane detection method is introduced where the availability of 3D information allows the separation between the road and the obstacle features. Consequently, The lane can be modeled as a 3D surface and the prediction of its current parameters is performed using past information and vehicle dynamics. 
\section{Lane ID Estimation from monocular images}
\subsection{Dual Left-Right Convention}
\begin{figure}
\centering
\includegraphics[width=0.49\textwidth]{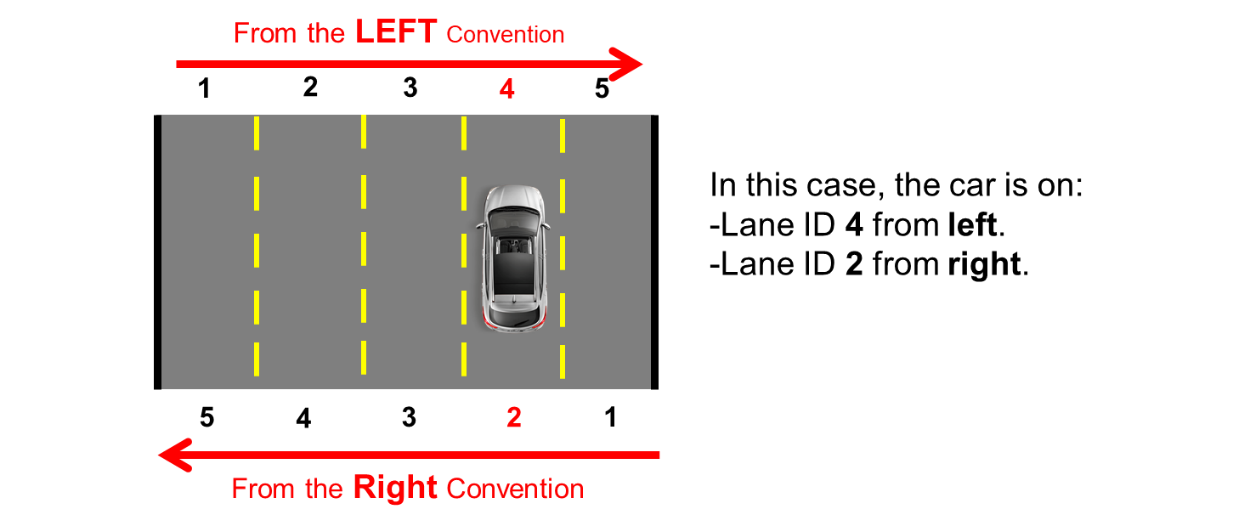}
\caption{Dual left-right convention for lane ID estimation} \label{dualconv}
\end{figure}

%A countless number of People and goods are transported daily to their destinations through the widely ramified network of roads going through each country. To guarantee safe and efficient traffic, boundaries and markings on roadways need to be clearly defined and made easily visible to travelers according to the traffic rules adopted by the country. Despite substantial differences in these rules, one important shared agreement is globally adopted concerning the fact that lane markings are there to delineate the drivable region of the road within which a single file of vehicles is supposed to cruise. In fact, roadways are generally divided into multiple corridors to define and organize traffic flow according driving direction and speed of traveling vehicles.  
In the context of autonomous driving, a reliable perception and interpretation of traffic rules in general and roadway lanes in particular is necessary for the vehicles to ensure safe traveling. It is of high importance that autonomous cars are reliably aware of the road structure to be able to make proper driving decisions. For this work, we define the problem of lane ID estimation as the task of determining the relative number or identifier ID of the current lane driven based only on the content of a scene captured by a monocular camera. Unlike the standard lane estimation problematic generally defined as the task of figuring out the semantic, topological and geometric properties of the driven road, our present work comes as the answer to a very specific question concerning the number (ID) of the currently driven lane out of many surrounding the vehicle. This ID is defined with respect to a pre-fixed reference point, could be the left or the right border of the road. Two conventions can be used to define the corresponding lane ID, where left ID $\delta_l$ is calculated with respect to the left road side and $\delta_r$ with respect to the right one. These two conventions are related as follows:
\begin{equation} 
\delta_r = L_c - \delta_l + 1 
\end{equation} 
with $L_c$ being the lane count (total number of lanes in the image), $\delta_l$ the driven lane ID using the left convention and $\delta_r$ the right one. An example of our double convention definition could be seen in Fig.~\ref{dualconv}.

The motivation behind using this double scheme is to enforce more constraints on the estimated output of the network during the training phase. In fact, the model is supposed to deliver three classification vectors for left and right estimated lane IDs together with lane count. The enforced redundancy by using two conventions improves the likelihood that both or at least one of the ID estimates is correct. We approach the task from two different point of views, namely left and right sides of the road and get profit from this cooperation using always the most reliable side that offers more visual information and better features. This could be the case for different challenging situations such as strongly occluded vision or cluttered scenes. We also assume that estimating the relative lane ID with respect to the closest road side is more reliable since features and landmarks on the border are richer and easier to learn for the model. Hence, the more visual information the camera captures from the road side, the more accurate and reliable the network estimation is likely to be.   

\subsection{Model Architecture}

\begin{figure}
\centering
\includegraphics[width=0.51\textwidth]{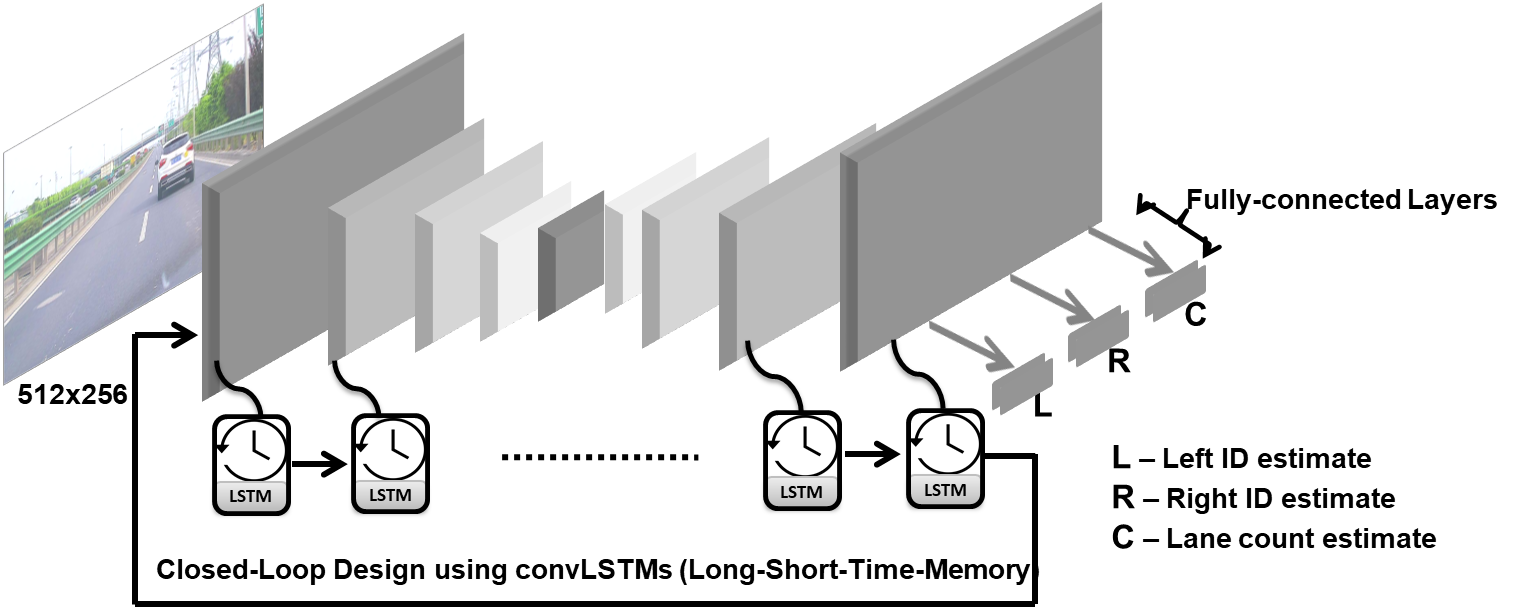}
\caption{The "Moka-convLSTM" architecture used for relative lane ID estimation} \label{moka}
\end{figure}

As the backbone of the present solution, we propose an in-house designed architecture for the neural network used to perform relative lane ID estimation inspired by the model suggested in~\cite{halfaoui2016cnn}. 
Our new architecture called "Moka-convLSTM" shown in Fig.~\ref{moka} is composed of an encoder part which gradually extracts high level features by down-sampling the image/feature maps, together with a decoder part which enables the recovery of the full resolution. Long-range dense links are used to connect both parts which improves the quality of the feature maps during both contractive and refinement stages. In fact, multiple feature maps representing different high level abstractions from previous stages are combined at each resolution level to obtain wide range of stacked features that are passed accordingly to the next layers using the same strategy. At the end, three independent full-connected blocks are linked to the final layer of the decoder in order to generate the required classification vectors corresponding to the left, right and lane count estimates. As a further extension, our network is equipped with ConvLSTM cells \cite{xingjian2015convolutional} to guarantee recurrence necessary to process sequences. These cells are interposed at each level in order to enable the capturing of the temporal dimension. Obviously, each one is constructed out of a memory cell $c_t$ accumulating state information over the given input sequence, a forget gate $f_t$ responsible for deciding upon how much information from the past cells should be retained and input and output gates $i_t$ and $o_t$ respectively responsible for receiving and emitting information. The main advantage of convLSTM is the ability of holding information on previous data introduced to the network and using it to make decisions about the current input or forecasting future states. In other words, they help the model remember and take into consideration distant occurrences from the past into the final output.
%Furthermore, few improvements are brought to the design of the three final blocks. These aim to guarantee a smooth transition from the last decoder layer to each fully connected block. In this step, the network needs to compress a $1 \times N \times H \times W$ tensor into $1 \times N$ vector over gradual intermediate steps meaningfully compressing the content instead of abrupt compression, where we directly go down with size from the two-dimensional original resolution to the final one. The proposed gradual compression is detailed in Fig.~\ref{bottelneck} where H and W are Height and Width of the original input image and $N$ represent the classification cardinality. In other words, it is the pre-defined maximum number of possible available lanes, in our case 8 lanes (we consider only roads with up to 8 lanes). 
%\begin{figure}[!h]
%\centering
%\includegraphics[width=0.33\textwidth]{images/bottelneck.png}
%\caption{The "Bottelneck" improvement in the proposed architecture } \label{bottelneck}
%\end{figure}

%
%
%
\subsection{Training Setup}
The proposed model was trained end-to-end to perform relative lane ID estimation on a set consisting of 
244 sequence of 2500 image each ($\sim$600k images). The data that cannot be disclosed due to legal and security restrictions was exclusively recorded in the Chinese city of Shanghai at different dates, seasons, weather conditions, day times and routes over a span of time from June 2018 until February 2019. For the labeling, a semi-automatic strategy was used to generate proper lane count and lane ID labels for each image following our previously detailed double convention scheme based on a high-accuracy GPS device and a pre-stored high-definition map of the city.  \\   
For training, we employ a custom version of Pytorch~\cite{pyt} and make use of the Adam optimizer~\cite{kingma2014adam}. We train the network up to 300K iterations with batches of size 2 containing sequences of consecutive frames of length 4 
and a learning rate $\lambda = 10^{-4}$ that is divided by 2 every 20k iterations starting from iteration 150k. We set momentum values to $\beta_1 = 0.9$ and $\beta_2 = 0.999$ with weight-decay $\delta = 10^{-4}$. 
%The number of output feature maps is set to $num_{output} = 32$ and the kernel size is $k = 4 $ for all the layers. The stride is a power of 2 depending on the wanted resolution and the padding is correspondingly chosen in respect to the used stride and kernel size. These parameters were presented in~\cite{halfaoui2016cnn} and empirically verified for our current application. 
Input images were introduced in sequential form under the resolution $H \times W = 256 \times 512$ and underwent heavy random augmentations performed on the fly...\\
Our proposed network performs a classification task to estimate the lane count, left and right IDs out of a total number of $N = 8$ classes . The proposed cost function for training the network is composed of three terms summarized as follows:
\begin{enumerate}
\item $l_{ce}$: A standard cross-entropy loss term for performing multi-class classification for each output defined as:
\begin{equation}   
l_{ce}(\underline{x},\underline{y}) = - \sum_{i}^{N} y_i \cdot log(x_i) 
\end{equation}  
with $\underline{x}$ is the estimated probability classification vector and $\underline{y}$ the corresponding ground-truth label written in one-hot encoding form as well, valid for all three output entities and their corresponding ground-truth labels.

\item $l_{ad}$: An adaptive penalty term that gives more weight to the smallest ID between the right and left estimates, which comes in line with our previous assumption that the nearest road side would be more suitable to use as reference point to define the lane ID. 
\begin{equation}  
l_{ad}(z) = 1 + e^{ -5 \cdot z}
\end{equation}  
with z the predicted scalar value for left $\widehat{\delta}_l$ or the right $\widehat{\delta}_r$ estimates.

\item $l_{cn}$: A triangular constraint enforcing the linearity dependence between the three predicted scalar outputs: right $\widehat{\delta}_r $, left $\widehat{\delta}_r$ and lane count $\widehat{L}_c$ estimated by the network and expected to fulfill the mathematical relation:
\begin{equation} 
 l_{cn} = \widehat{\delta}_r - \widehat{L}_c + \widehat{\delta}_l - 1 \mbeq 0 
\end{equation} 

\end{enumerate}

To sum up, the proposed final training cost function is defined as:

\begin{equation} 
 Loss = l_{ad}(\widehat{\delta}_l) \cdot l_{ce}(\underline{\widehat{\delta}_l}, \underline{{\delta}_l}) + l_{ad}(\widehat{\delta}_r) \cdot l_{ce}(\underline{\widehat{\delta}_r}, \underline{{\delta}_r}) + 
l_{ce}(\underline{\widehat{L}_c}, \underline{L_c}) + l_{cn} 
 \end{equation}

with $\delta_l, \delta_r, L_c$ are the scalar ground-truth labels to left, right (IDs) and lane count respectively. $\underline{\widehat{\delta}_r}, \underline{\widehat{\delta}_l}, \underline{\widehat{L}_c}$  are the estimated probability classification vectors by the network, $\underline{\delta_r}, \underline{\delta_l}, \underline{L_c}$ are the corresponding ground-truth labels in one-hot encoding form and $\widehat{\delta}_l, \widehat{\delta}_r, \widehat{L}_c$ are the final scalar estimates for IDs and lane count. 
\section{Experiments}
\subsection{Brightness Adjustment For Recurrent Models}
We propose a pre-processing step to adjust the brightness level of input images. Obviously, images of a single input sequence need to depict reasonable brightness levels without strong fluctuations where features are easily extractable and learnable. Therefore, adjusting the brightness domain in which the recurrent network operates proves necessary as the training set presents a rich panoply of weather conditions, day times and seasons, which in turn causes considerable illumination variations among the training images. Particularly, we notice that any significant abrupt change in terms of brightness for images of the same input sequence is tightly correlated with a decrease in our lane ID estimation accuracy, as past information tracked through convLSTM cells contributes significantly to shaping the decision of the network. So if the model is presented with a sequence with considerable brightness fluctuations, the convLSTM cells will not get profit from tracking past information over time, but would rather suffer from it as current and past internal states present meaningful differences.  \\
As a measure for enforcing consistency, we first keep track of the running average perceived brightness from the input images belonging to the same sequence. Then, we process each coming image in the deployment phase before introducing it to the model according to how its depicted brightness compares to the average tracked value. Specifically, if the perceived brightness of this current frame is below the tracked value then we adjust its brightness via a linear transformation of pixel intensities inspired by gamma correction. 
%This particular case is widely common and could be faced in the exact same example previously mentioned. We refer here to the already discussed case where GPS devices fail spectacularly in the challenging situation of entering a tunnel or engaging in a down-town drive where luminosity gets affected by higher buildings, bridges or any shadow causing agent.
As an example, let's assume we have a car driving on a normal highway in clear conditions then a tunnel emerges. If image I, the currently processed image, is the first frame captured inside the tunnel, then it would depict totally different brightness properties compared to past images, as the lighting conditions had changed abruptly. Let's assume that I could be divided into three color channels red $R_I$, blue $B_I$ and green $G_I$ with $b_I$ its perceived brightness and $\tilde{b}$ the running average tracked brightness value over all images captured outside the tunnel. We can first calculate the adjusting factor $\alpha$ as:
\begin{equation} 
 \alpha = \frac{\tilde{b}}{b_I} 
 \end{equation} 
Then, we use this factor to generate new color channels for the image I by combining three new channel matrices $R'_I, B'_I$ and $G'_I$ produced as follows with pixel location defined by the tuple $(x,y)$:
\begin{equation} 
\begin{array}{c} 
  R'_I(x,y) = min(255, \quad \alpha \cdot R_I(x,y)) \\
  B'_I(x,y) = min(255, \quad \alpha \cdot B_I(x,y)) \\
  G'_I(x,y) = min(255, \quad  \alpha \cdot G_I(x,y))
 \end{array}
\end{equation}
After reconstructing the brightness adjusted image out of the newly generated color channels, we introduce it finally to the pre-trained model to perform relative lane ID estimation. 
%Obviously, our model won't experience a big difference in input's properties and it can smoothly perform the prediction on the modified image, that presents now similar brightness level in comparison to the previously processed part of the sequence. Eliminating this disturbing factor for the pre-trained model, guarantees consistent input data brightness domain. This subsequently improves the estimation accuracy as convLSTM cells gain full potential from the past information regardless the strongly varying lighting conditions all along the processed sequence.  

\subsection{Decision Module For Final Used Convention}
In the deployment phase, our model delivers simultaneously two ID estimates $\delta_l$ and $\delta_r$ according to left and right conventions for each input image together with the lane count. Hence, an additional decision step should be considered to pick up a single final output out of these two candidates. Even if both estimates are correct, only a single ID should be considered otherwise a confusion might arise for the user. That is why, specifying the final estimated lane ID must be always done in reference to the adopted convention. Many possible ways could be harnessed to make the decision. However, we opt for a specific technique we define as the entropy-based decision (we borrow the "entropy" concept from physics to describe how dispersed is the distribution of a certain set of elements).
%and we show it in Fig.~\ref{decision}
This decision process could be summarized as follows: \\
After getting the output classification vectors delivered by the pre-trained model, we consider each one separately: We first select the maximum probability value out of the N elements of each one. Then, we calculate the mean value out of all elements of the vector. Finally, we subtract the resulting mean out of the pre-selected maximum and we end up with values describing roughly the distribution of elements in each vector. These are considered to make the decision about the final output lane ID based on their comparison and the one with the higher value should be adopted as the final estimate.\\ 
\begin{figure}
\centering
\includegraphics[width=0.45\textwidth]{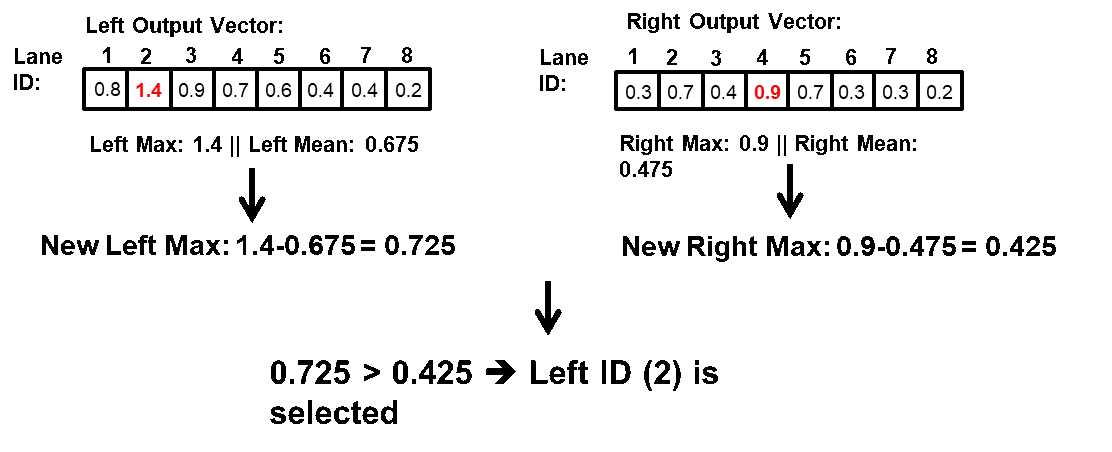}
\caption{The decision module for final adopted convention} \label{decision}
\end{figure}
Furthermore, a penalizing term against fluctuations between consecutive estimates for adjacent input images is additionally introduced. The idea is to enforce a smooth estimation between consecutive IDs by avoiding unreasonable jumps in the predicted lanes between consecutive frames. Obviously, a vehicle cannot abruptly move from the first to the seventh lane, for example, within the very short span of time separating the capturing of the two images. Therefore before applying the comparison between the previously discussed entropy values, we weight them accordingly with a corresponding factor P specific to each convention, which can be calculated as follows:  

\begin{equation} 
 P = \frac{1}{1 + |O_t - O_{t-1}|} 
\end{equation} 
with $O_t$ is the final output at the current time $t$ and $O_{t-1}$ is the output of the previous frame $t-1$ using the same convention.

\subsection{Evaluation}

\begin{table*}
\caption{Numerical performance analysis for custom architectures Alexnet, VGG, Resnet, Densenet, ResNext, Shufflenet, Mobilenet and different versions of our Moka model (basic, convLSTM, standard LSTM) on test set}\label{tab1}
\centering
\scalebox{0.62}{
%\begin{tabular}{|l||c|c|c|c|c|c|c|}
 \begin{tabular}{| >{\centering\arraybackslash}m{1.2in} ||>{\centering\arraybackslash}m{1.01in} | >{\centering\arraybackslash}m{0.95in} | >{\centering\arraybackslash}m{0.9 in} | >{\centering\arraybackslash}m{0.99in} | >{\centering\arraybackslash}m{0.6 in} | >{\centering\arraybackslash}m{.6in} | >{\centering\arraybackslash}m{.8in} |}
\hline
{\bf Model} &  {\bf Raw combined Accuracy (best case)} (\%)  &  {\bf Final combined Accuracy (real case} (\%) & {\bf Accuracy Left ID only} (\%) & {\bf Accuracy Right ID only} (\%)  & {\bf Time} (ms) & {\bf Size} (MB) & {\bf Parameters} (Millions)\\ 
\hline \hline
VGG19 &    97.04  & 93.65  & 89.47  & 84.56  &  214 & 532 & 139.6 \\ 
VGG16 &   \textbf{97.50 } &  \textbf{94.73 } & \textbf{90.48 } & \textbf{85.64}   & 210  &  512 & 134.4 \\ 
Alexnet &   61.31  & 50.41  & 50.03  & 37.02 &  190 & 217 & 57.1 \\ 
Resnet &    72.91  & 58.02  & 55.22 & 33.37  & 173 & 90 & 23.1 \\ 
Densenet &   95.25  & 89.93  & 86.52  & 83.18  & 88 & 27 & 6.9  \\ 
%Squeezenet &   91.33  & 85.23  & 78.77  & 69.05  & \textbf{16} & \textbf{5} & \textbf{1.3} \\ 
Mobilenet &   89.17  & 78.34  & 68.09  & 64.16  & 26 & 9 & 2.2 \\
Shufflenet &   85.52  & 74.70  & 72.12  & 55.44  & 56 & 21 & 5.3 \\
ResNeXt    &   96.58  & 92.41  & 89.35  & 80.03  & 115 & 88 & 23.0 \\  
MOKA-basic &   92.75  & 86.23  & 82.14  & 78.47  & \textbf{18} & \textbf{7}  & \textbf{1.7} \\  \hline \hline
MOKA-StdLSTM &  96.17  & 92.27  & 89.22  & 79.56  &  24 & 10 & 1.8 \\
MOKA-convLSTM &  96.75  & 93.94  & 89.68  & 80.37   & 27 & 11 & 2.4 \\  
\hline
\end{tabular}
}
\end{table*}

\begin{table*}
\caption{Performance analyis using different brightness thresholds for pre-processing before applying the various models on test set}\label{tab2}
\centering
\scalebox{0.64}{
%\begin{tabular}{|p{3cm}|p{2cm}||p{2cm}||p{1.3cm}|p{1.3cm}|p{1.3cm}|p{1.3cm}|}
 \begin{tabular}{| >{\centering\arraybackslash}m{1.2in} |>{\centering\arraybackslash}m{0.8in} || >{\centering\arraybackslash}m{0.9in} | >{\centering\arraybackslash}m{0.7 in} | >{\centering\arraybackslash}m{.7in} |>{\centering\arraybackslash}m{.7in}  | >{\centering\arraybackslash}m{.7in} |}   
\hline 
\textbf{Model} & \textbf{Accuracy (\%)} & \textbf{No Brightness Adjustment} &  \textbf{B = 100} & \textbf{B = 130} & \textbf{B = 150} & \textbf{B = 170}\\
\hline \hline
\multirow{2}{*}{VGG19} & Raw &  97.06 \% &  97.02 \% & 96.96 \% & 96.76 \% & 96.39 \%     \\ 
                       & Final &  93.65 \% & 93.56 \% & 93.55 \% & 93.20 \% & 92.81 \%  \\ \hline
\multirow{2}{*}{VGG16} & Raw & \underline{97.50} \% & \underline{97.42} \% & 97.40 \% & 97.26 \% &  96.95 \%  \\ 
                       & Final &  \underline{94.73} \% & \underline{94.66} \% &  94.55 \% & 94.37 \% & 93.86 \%   \\ \hline
\multirow{2}{*}{Alexnet} & Raw &  61.31 \% & 61.33 \% & 61.31 \% & 61.31 \% & 61.31 \%  \\ %\cline{2-7}
                      	 & Final &  50.41 \% & 50.41 \% & 50.41 \% & 50.41 \% & 50.41 \%  \\ \hline
\multirow{2}{*}{Resnet} & Raw &  72.91 \% &   72.03 \% & 74.17 \% & 71.22 \% & 64.05 \%  \\  
                      	& Final &  58.02 \% & 57.41 \% & 57.83 \% & 54.94 \% & 42.12 \%   \\ \hline                      
\multirow{2}{*}{Densenet} & Raw &  95.25 \% & 95.23 \% & 94.88 \% & 94.72 \% & 94.47 \%  \\ 
                      	  & Final &  89.93 \% & 89.27 \% & 89.81 \% & 89.38 \% & 88.92 \%    \\ \hline                          \multirow{2}{*}{Mobilenet} & Raw & 87.19 \%  &  87.18 \% & 87.22 \% & 87.30 \% & 87.23 \%   \\  
                      	 	& Final &  78.24 \% & 78.25 \% & 78.22 \% & 78.34 \% & 78.04 \%  \\ \hline                      	\multirow{2}{*}{Shufflenet} & Raw & 85.52 \%  &  85.52 \% & 85.46 \% & 85.27 \% & 84.94 \%   \\ 
                      	 	& Final &  74.70 \% & 74.70 \% & 74.63 \% & 74.38 \% & 73.89 \%  \\ \hline                      	\multirow{2}{*}{ResNeXt}   & Raw & 96.58 \%  &  96.58 \% & 96.51 \% & 96.45 \% & 96.25 \%   \\  
                      	 	& Final &  92.41 \% & 92.31 \% & 92.36 \% & 92.39 \% & 92.18 \%  \\ \hline                                         \multirow{2}{*}{MOKA-basic} & Raw & 92,75 \% & 87.12 \% & 93.34 \% & 90.82  \% & 88.31 \%    \\ 
                       		& Final &  86.23 \% &   76.61\% &  87.80 \% &  86.64 \% & 84.54 \%    \\ \hline
\multirow{2}{*}{MOKA-StdLSTM} & Raw &  96.17 \% & 89.62 \% & 97.48 \% & 97.32 \% & 96.66 \%  \\ 
                      	 	  & Final & 92.27 \% & 81.68 \% & 94.82 \% & 94.14 \% & 93.22 \%   \\ \hline
\multirow{2}{*}{MOKA-convLSTM} & Raw &  96.75 \% & 90.74 \% & \textbf{ \underline {98.71} \%} & \underline{98.41} \% &  \underline{98.02} \% \\ 
                       		   & Final &  93.94 \% & 85.94 \% & \textbf{ \underline {95.36} \%} & \underline{95.14} \% & \underline{94.61} \%  \\                       
\hline
\end{tabular}}
\end{table*}
%\multirow{2}{*}{Squeezenet} & Raw &  91.33 \% & 91.31 \% & 91.14 \% & 90.78 \% & 90.17 \%    \\  
%                      	 	& Final &  85.23 \% & 85.17 \% & 85.02 \% & 84.89 \% & 83.65 \%  \\ \hline                                        

\begin{figure*}
\centering
\includegraphics[width=0.73\textwidth]{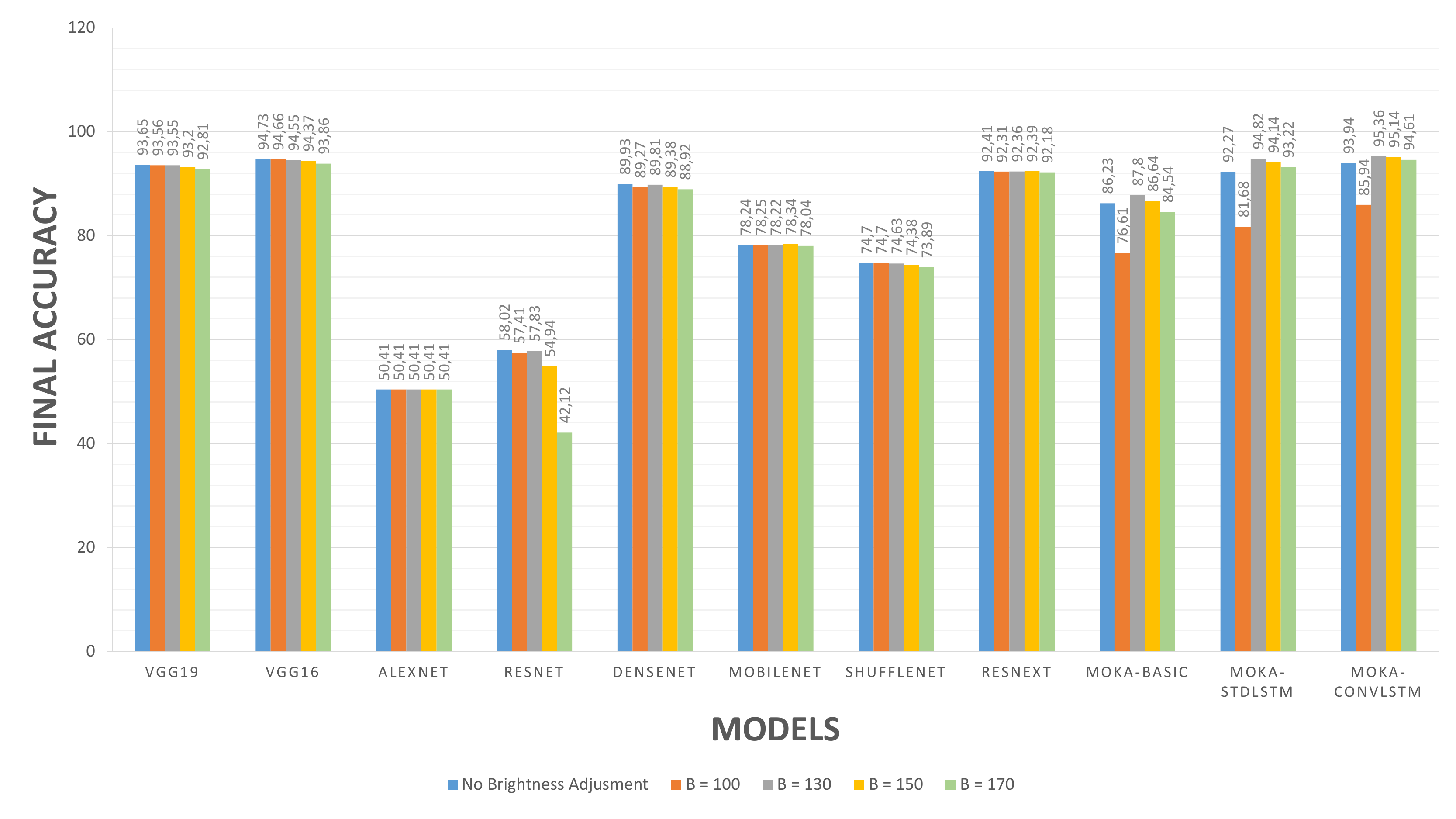}
\caption{ Final accuracies using different pre-processing setups for brightness adjustment } \label{final_acc}
\end{figure*}

For evaluation, we run a battery of experiments to assess the proposed approach. As the nature of the targeted application is quite specific, the performed literature research didn't reveal other methods aiming to solve the precise task of visual end-to-end lane ID estimation. Generally, the community is interested in a larger scope including applications such as lane segmentation~\cite{oliveira2016efficient,jang2018road,meyer2018deep}, estimation~\cite{kim2017end,gurghian2016deeplanes,rabe2016ego}, prediction~\cite{tang2018lane,son2015predictive} or detection~\cite{li2016deep,lee2017vpgnet,niu2016robust,jung2015efficient}.  
Although these methods could be leveraged to solve the task at hand, this would however require additional intermediate steps and further processing to estimate the current lane ID. Since we are interested in an efficient solution that is aimed to be a building block of a large processing chain for high-complexity applications such as mapping, localization and path planing, a complex and sophisticated approach that is expensive in time and resources would be unfit for these purposes. That is why, we avoid considering complex methods and focus rather on the important aspects that can make the expected solution suitable to the requirements of the usecase (efficiency, realtime usage and accuracy). We analyse the performance of some popular custom architectures for lane ID estimation from monocular images. These include Alexnet ~\cite{krizhevsky2014weird}, VGG~\cite{simonyan2014deep}, Resnet~\cite{He_2016} and Densenet~\cite{Huang_2017}... For the sake of consistency, all models were trained using the previously detailed setup and tested on a new set including 163 unfamiliar sequences with 2500 image each ($\sim$400k).\\
\begin{table*}[!h]
\caption{Performance comparison (final accuracy) using different decision criteria for the choice between left and right conventions. Considering left and right classification vectors, we compare Max: Maximum values, Max-M: Maximums - means of the corresponding vectors, E: Entropy values, Max-E: Maximums - entropys, Z-score: Statistical standard scores. }\label{tab3}
\centering
\scalebox{0.69}{
%\begin{tabular}{|l||l|l|l|l|l|}
\begin{tabular}{| >{\centering\arraybackslash}m{1.6in} ||>{\centering\arraybackslash}m{0.8in} | >{\centering\arraybackslash}m{0.7in} | >{\centering\arraybackslash}m{0.7 in} | >{\centering\arraybackslash}m{.7in} | >{\centering\arraybackslash}m{.7in} |}
\hline
 \textbf{Model} &  \textbf{Max} & \textbf{Max-M} & \textbf{E} & \textbf{Max-E} & \textbf{Z-score} \\
\hline \hline
VGG19 &  93.43 \% & 93.65 \% & 81.86 \% & \underline{93.37} \% &  94.22 \% \\
VGG16 &  94.47 \% & 94.73 \% & \underline{83.23} \% &  92.27 \% &  94.95 \% \\ 
Alexnet &  49.84 \% & 50.41 \% & 47.02 \% & 49.14 \% &  49.83 \% \\
Resnet & 40.67 \% & 58.02 \% & 30.29 \% &  49.89 \% &  43.93 \%  \\
Densenet &  89.99 \% & 89.93 \% & 73.37 \% &  89.11 \% &  90.24 \% \\ 
%Squeezenet &   78.58 \% & 85.23 \% & 87.49 \% &  65.23 \% &  84.91 \%  \\
Mobilenet &   78.06 \% & 78.24 \% & 57.96 \% &  75.24 \% &  77.44 \%  \\
Shufflenet &   74.35 \% & 74.70 \% & 53.80 \% &  74.68 \% &  74.24 \%  \\
ResNext &   92.39 \% & 92.41 \% & 81.76 \% &  88.61 \% &  91.79 \%  \\
MOKA-basic &  86.11     \% & 86.23 \% &  67.37  \% &  70.56 \% &  82.35 \%\\ 
MOKA-StdLSTM-B130 &  92.21 \% & 94.82 \% & 76.04 \% &  89.88 \% &  92.33 \% \\ 
MOKA-convLSTM-B130 & \textbf{\underline{95.47}} \% & \underline{95.36} \% & 83.09 \% &  84.21 \% &  \underline{95.02} \%  \\
\hline
\end{tabular}}
\end{table*}
\begin{figure*}[!h]
  \centering
  \scalebox{0.36}{
  \begin{tabular} { p{.9\textwidth}  p{.9\textwidth}   } 

\includegraphics[width=.88\textwidth, height =.44\textwidth]{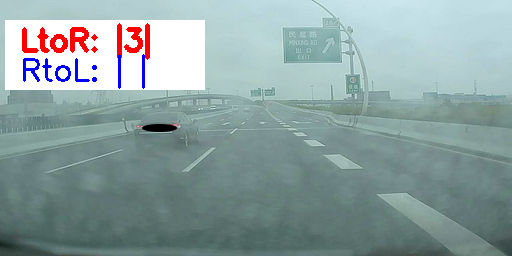} &
\includegraphics[width=.88\textwidth, height =.44\textwidth]{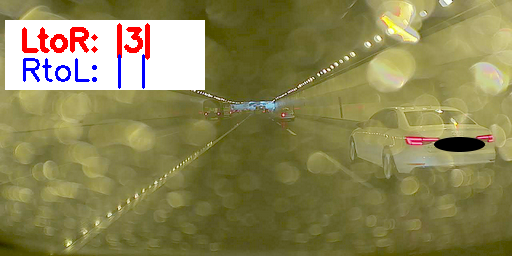}\\ \\

\includegraphics[width=.88\textwidth, height =.44\textwidth]{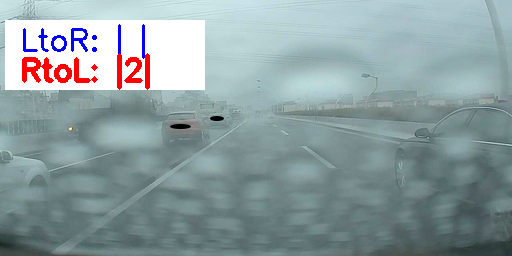} &
\includegraphics[width=.88\textwidth, height =.44\textwidth]{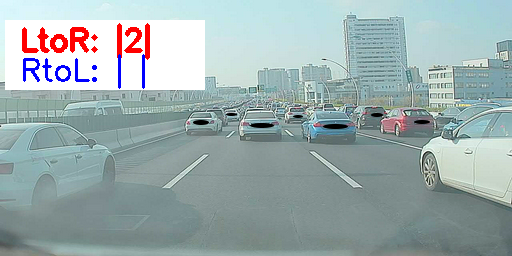} \\ \\ 

\end{tabular}
}
\caption{Visual result samples using best performing model Moka-convLSTM-B130 (LtoR: left convention, RtoL: right convention)}
\label{fig:e}
\end{figure*}
A numerical comparison between the architectures in terms of performance, required runtime for processing a single image on an NVIDIA Geforce GTX TITAN X, memory size and parameters number is shown in table~\ref{tab1}.
Results indicate that our dual convention scheme offers meaningful improvement against considering left and right accuracies seperately. This is valid for all models and their corresponding raw accuracy values (the upper bound case where we use always the best out of the two conventions, i.e. at least one ID is correct), as well as for the final accuracy which is calculated after applying a decision module to choose a single output out of both candidates (might not be always correct choice).  
Obviously, reaching high accuracy levels comes at cost of the complexity aspect. Hence, it is then expected that both models with the highest complexity levels, namely Vgg16 and Vgg19, are the best performing in terms of raw as well as final accuracy. Our basic Moka model (without recurrence) designed to fulfill the tight time and memory requirements of our buisness usecase performs quite decently considering its limited number of parameters with 92.75 \% raw accuracy and 86.23\% final one. The question that arised here was how to improve the performance of our basic Moka while keeping its complexity low?\\
To answer this, we decided to rely on the temporal dimension of input image sequences instead of stacking additional layers atop of the current architecture. Two alternatives were considered to improve the performance without loosing too much on complexity. First, we extended the fully connected blocks after the decoder part with standard LSTM cells to track past estimates, where inputs are the one-dimensional classification vectorst. This version we named Moka-stdLSTM. Second, we proposed the Moka-convLSTM version with interposed memory cells at each convolutional layer to keep track of the slightest details aggregated from previous runs. Unlike the standard case, the incoming cell inputs are two-dimensional (2D) convolved feature maps. Both modified versions improved significantly upon the original without sacrifying the important properties of the basic Moka model. Although slightly ouperformed, they reached the accuracy range of the best networks with signficant advantage in terms of required runtime and complexity as shown in table~\ref{tab1}.\\
%After getting close to the best range thanks to considering the temporal dimension, just a small margin was left to our lightweight recurrent models to outperform these networks. 
As a further improvement measure, we proposed, as previously discussed, the additional pre-processing step to enforce brightness level consistency. A thourough examination of different brightness thresholds was performed during experiments but for the sake of simplicity we only show the results with few chosen values in table~\ref{tab2}. Obviously, this measure has way more effect on recurrent models considering the temporal dimension than custom networks trained on single images as depicted in table~\ref{tab2} and Fig~\ref{final_acc}. Despite the additional time overhead needed for each incoming image, the low-complexity recurrent models still fulfill the required runtime constraints and are now clearly ahead of the VGG models in terms of accuracy with MOKA-convLSTM-B130 as new best performer achieving 98.41\% and 95.36\% for raw and final accuracy, respectively. \\
Obviously, the difference between raw and final accuracies for all models is a sign that the decision step we use to choose between left and right conventions can be improved. There is still a potential gap to close between the final reached accuracy and the best performance that we can achieve if we fully get profit from the model estimation for both conventions.  As a final test, we consider several decision criteria used in combination with all models to thouroughly explore the impact of this step on final accuracies. The goal is to find the best decision strategy to minimize this gap. \\
The comparison shown in table~\ref{tab3} shows that using different criteria has different impact on performance when used in combination with varying models. However, we observe that the recurrent model MOKA-convLSTM-B130 is still best performer in three cases out of five and more imporantly the best overall performer despite a sustained difference between raw accuracy 98.71\% and the new final accuracy value 95.47 \%. 

\section{Conclusion}
In the context of autonomous driving, there is a growing consensus that the more boundaries we push towards full autonomy, the more specific and challenging the technical issues to be addressed become. 
In this work, we provide a solution to the particular task of lane ID estimation that can be beneficial to several applications such as mapping and localization, path planing and safe path estimation \\
Our novel end-to-end low-complexity solution relies only on monocular visual information and harnesses the temporal dimension inherent to the input sequences to yield precise lane ID estimation for real-time requirements using a convLSTM-based network.

\bibliographystyle{named}
\bibliography{ijcai20}

\begin{thebibliography}{}

\bibitem[\protect\citeauthoryear{Aly \bgroup \em et al.\egroup
  }{2015}]{lanequest}
Heba Aly, Anas Basalamah, and Moustafa Youssef.
\newblock Lanequest: An accurate and energy-efficient lane detection system.
\newblock In {\em 2015 IEEE International Conference on Pervasive Computing and
  Communications (PerCom)}, pages 163--171. IEEE, 2015.

\bibitem[\protect\citeauthoryear{Cui \bgroup \em et al.\egroup
  }{2015}]{stereov}
Dixiao Cui, Jianru Xue, and Nanning Zheng.
\newblock Real-time global localization of robotic cars in lane level via lane
  marking detection and shape registration.
\newblock {\em IEEE Transactions on Intelligent Transportation Systems},
  17(4):1039--1050, 2015.

\bibitem[\protect\citeauthoryear{Dao \bgroup \em et al.\egroup
  }{2007}]{intvec_com}
Thanh-Son Dao, Keith Yu~Kit Leung, Christopher~Michael Clark, and Jan~Paul
  Huissoon.
\newblock Markov-based lane positioning using intervehicle communication.
\newblock {\em IEEE Transactions on Intelligent Transportation Systems},
  8(4):641--650, 2007.

\bibitem[\protect\citeauthoryear{Gurghian \bgroup \em et al.\egroup
  }{2016}]{gurghian2016deeplanes}
Alexandru Gurghian, Tejaswi Koduri, Smita~V Bailur, Kyle~J Carey, and Vidya~N
  Murali.
\newblock Deeplanes: End-to-end lane position estimation using deep neural
  networksa.
\newblock In {\em Proceedings of the IEEE Conference on Computer Vision and
  Pattern Recognition Workshops}, pages 38--45, 2016.

\bibitem[\protect\citeauthoryear{Halfaoui \bgroup \em et al.\egroup
  }{2016}]{halfaoui2016cnn}
Ibrahim Halfaoui, Fahd Bouzaraa, and Onay Urfalioglu.
\newblock Cnn-based initial background estimation.
\newblock In {\em 2016 23rd International Conference on Pattern Recognition
  (ICPR)}, pages 101--106. IEEE, 2016.

\bibitem[\protect\citeauthoryear{He \bgroup \em et al.\egroup }{2016}]{He_2016}
Kaiming He, Xiangyu Zhang, Shaoqing Ren, and Jian Sun.
\newblock Deep residual learning for image recognition.
\newblock {\em 2016 IEEE Conference on Computer Vision and Pattern Recognition
  (CVPR)}, Jun 2016.

\bibitem[\protect\citeauthoryear{Huang \bgroup \em et al.\egroup
  }{2017}]{Huang_2017}
Gao Huang, Zhuang Liu, Laurens van~der Maaten, and Kilian~Q. Weinberger.
\newblock Densely connected convolutional networks.
\newblock {\em 2017 IEEE Conference on Computer Vision and Pattern Recognition
  (CVPR)}, Jul 2017.

\bibitem[\protect\citeauthoryear{Jang \bgroup \em et al.\egroup
  }{2018}]{jang2018road}
Wonje Jang, Jhonghyun An, Sangyun Lee, Minho Cho, Myungki Sun, and Euntai Kim.
\newblock Road lane semantic segmentation for high definition map.
\newblock In {\em 2018 IEEE Intelligent Vehicles Symposium (IV)}, pages
  1001--1006. IEEE, 2018.

\bibitem[\protect\citeauthoryear{Jung \bgroup \em et al.\egroup
  }{2015}]{jung2015efficient}
Soonhong Jung, Junsic Youn, and Sanghoon Sull.
\newblock Efficient lane detection based on spatiotemporal images.
\newblock {\em IEEE Transactions on Intelligent Transportation Systems},
  17(1):289--295, 2015.

\bibitem[\protect\citeauthoryear{Kim and Park}{2017}]{kim2017end}
Jiman Kim and Chanjong Park.
\newblock End-to-end ego lane estimation based on sequential transfer learning
  for self-driving cars.
\newblock In {\em Proceedings of the IEEE Conference on Computer Vision and
  Pattern Recognition Workshops}, pages 30--38, 2017.

\bibitem[\protect\citeauthoryear{Kingma and Ba}{2014}]{kingma2014adam}
Diederik~P Kingma and Jimmy Ba.
\newblock Adam: A method for stochastic optimization.
\newblock {\em arXiv preprint arXiv:1412.6980}, 2014.

\bibitem[\protect\citeauthoryear{Knoop \bgroup \em et al.\egroup
  }{2017}]{gpsppp}
Victor~L Knoop, Peter~F de~Bakker, Christian~CJM Tiberius, and Bart van Arem.
\newblock Lane determination with gps precise point positioning.
\newblock {\em IEEE Transactions on Intelligent Transportation Systems},
  18(9):2503--2513, 2017.

\bibitem[\protect\citeauthoryear{Krizhevsky}{2014}]{krizhevsky2014weird}
Alex Krizhevsky.
\newblock One weird trick for parallelizing convolutional neural networks,
  2014.

\bibitem[\protect\citeauthoryear{Lee \bgroup \em et al.\egroup
  }{2017}]{lee2017vpgnet}
Seokju Lee, Junsik Kim, Jae Shin~Yoon, Seunghak Shin, Oleksandr Bailo, Namil
  Kim, Tae-Hee Lee, Hyun Seok~Hong, Seung-Hoon Han, and In~So~Kweon.
\newblock Vpgnet: Vanishing point guided network for lane and road marking
  detection and recognition.
\newblock In {\em Proceedings of the IEEE International Conference on Computer
  Vision}, pages 1947--1955, 2017.

\bibitem[\protect\citeauthoryear{Li \bgroup \em et al.\egroup
  }{2016}]{li2016deep}
Jun Li, Xue Mei, Danil Prokhorov, and Dacheng Tao.
\newblock Deep neural network for structural prediction and lane detection in
  traffic scene.
\newblock {\em IEEE transactions on neural networks and learning systems},
  28(3):690--703, 2016.

\bibitem[\protect\citeauthoryear{Meyer \bgroup \em et al.\egroup
  }{2018}]{meyer2018deep}
Annika Meyer, N~Ole Salscheider, Piotr~F Orzechowski, and Christoph Stiller.
\newblock Deep semantic lane segmentation for mapless driving.
\newblock In {\em 2018 IEEE/RSJ International Conference on Intelligent Robots
  and Systems (IROS)}, pages 869--875. IEEE, 2018.

\bibitem[\protect\citeauthoryear{Nedevschi \bgroup \em et al.\egroup
  }{2004}]{3dlane}
Sergiu Nedevschi, Rolf Schmidt, Thorsten Graf, Radu Danescu, Dan Frentiu,
  Tiberiu Marita, Florin Oniga, and Ciprian Pocol.
\newblock 3d lane detection system based on stereovision.
\newblock In {\em Proceedings. The 7th International IEEE Conference on
  Intelligent Transportation Systems (IEEE Cat. No. 04TH8749)}, pages 161--166.
  IEEE, 2004.

\bibitem[\protect\citeauthoryear{Niu \bgroup \em et al.\egroup
  }{2016}]{niu2016robust}
Jianwei Niu, Jie Lu, Mingliang Xu, Pei Lv, and Xiaoke Zhao.
\newblock Robust lane detection using two-stage feature extraction with curve
  fitting.
\newblock {\em Pattern Recognition}, 59:225--233, 2016.

\bibitem[\protect\citeauthoryear{Oliveira \bgroup \em et al.\egroup
  }{2016}]{oliveira2016efficient}
Gabriel~L Oliveira, Wolfram Burgard, and Thomas Brox.
\newblock Efficient deep models for monocular road segmentation.
\newblock In {\em 2016 IEEE/RSJ International Conference on Intelligent Robots
  and Systems (IROS)}, pages 4885--4891. IEEE, 2016.

\bibitem[\protect\citeauthoryear{pyt}{2019 accessed July 7 2019}]{pyt}
{\em Pytorch version 1.0}, 2019 (accessed July 7, 2019).

\bibitem[\protect\citeauthoryear{Rabe \bgroup \em et al.\egroup
  }{2016}]{rabe2016ego}
Johannes Rabe, Marc Necker, and Christoph Stiller.
\newblock Ego-lane estimation for lane-level navigation in urban scenarios.
\newblock In {\em 2016 IEEE Intelligent Vehicles Symposium (IV)}, pages
  896--901. IEEE, 2016.

\bibitem[\protect\citeauthoryear{Simonyan and
  Zisserman}{2014}]{simonyan2014deep}
Karen Simonyan and Andrew Zisserman.
\newblock Very deep convolutional networks for large-scale image recognition,
  2014.

\bibitem[\protect\citeauthoryear{Son \bgroup \em et al.\egroup
  }{2015}]{son2015predictive}
Young~Seop Son, Wonhee Kim, Seung-Hi Lee, and Chung~Choo Chung.
\newblock Predictive virtual lane method using relative motions between a
  vehicle and lanes.
\newblock {\em International Journal of Control, Automation and Systems},
  13(1):146--155, 2015.

\bibitem[\protect\citeauthoryear{Tang \bgroup \em et al.\egroup
  }{2018}]{tang2018lane}
Jinjun Tang, Fang Liu, Wenhui Zhang, Ruimin Ke, and Yajie Zou.
\newblock Lane-changes prediction based on adaptive fuzzy neural network.
\newblock {\em Expert Systems with Applications}, 91:452--463, 2018.

\bibitem[\protect\citeauthoryear{Xingjian \bgroup \em et al.\egroup
  }{2015}]{xingjian2015convolutional}
SHI Xingjian, Zhourong Chen, Hao Wang, Dit-Yan Yeung, Wai-Kin Wong, and
  Wang-chun Woo.
\newblock Convolutional lstm network: A machine learning approach for
  precipitation nowcasting.
\newblock In {\em Advances in neural information processing systems}, pages
  802--810, 2015.

\end{thebibliography}

\end{document}